\documentclass{article}
\usepackage{spconf,amsmath,graphicx}
\usepackage{amssymb}
\usepackage[pagebackref,breaklinks,colorlinks]{hyperref}
\usepackage{booktabs}
\usepackage{listings}
\usepackage[most]{tcolorbox}
\usepackage{enumitem}   
\usepackage{caption}    

\tcbset{
  prompt/.style={
    enhanced,
    colback=white,
    colframe=black,
    boxrule=0.8pt,
    arc=2mm,
    left=2mm,right=2mm,top=1.5mm,bottom=1.5mm,
    fonttitle=\bfseries
  }
}

\title{MedSAE: Dissecting MedCLIP Representations with Sparse Autoencoders}
\name{Riccardo Renzulli$^{\star}$ \qquad Colas Lepoutre$^{\dagger}$ \qquad Enrico Cassano$^{\star}$ \qquad Marco Grangetto$^{\star}$}

\address{$^{\star}$ University of Turin, Italy \\
$^{\dagger}$ École polytechnique, Palaiseau, France}

\begin{document}
\maketitle              
\begin{abstract}
Artificial intelligence in healthcare requires models that are accurate and interpretable. We advance mechanistic interpretability in medical vision by applying Medical Sparse Autoencoders (MedSAEs) to the latent space of MedCLIP, a vision-language model trained on chest radiographs and reports. To quantify interpretability, we propose an evaluation framework that combines correlation metrics, entropy analyses, and automated neuron naming via the MedGemma foundation model. Experiments on the CheXpert dataset show that MedSAE neurons achieve higher monosemanticity and interpretability than raw MedCLIP features. Our findings bridge high-performing medical AI and transparency, offering a scalable step toward clinically reliable representations. The source code supporting the findings of this study is  available at \url{https://github.com/EIDOSLAB/MedSAE}.
\end{abstract}


\section{Introduction}\label{sec:intro}
The emergence of Artificial Intelligence (AI) in healthcare systems is revolutionizing the way patients are diagnosed, treated, and monitored. Deep learning models have achieved remarkable results in various medical tasks \cite{LITJENS201760}. As the volume of medical data continues to grow, the size and complexity of neural network architectures have also increased. Self-supervised approaches now enable these models to train on diverse data sources (e.g., text, signals, images) with minimal reliance on human annotations, leading to the development of Multimodal Large Language Models (MLLMs).

Despite their advancements, MLLMs have notable limitations: they are computationally expensive and difficult to interpret. Mechanistic interpretability techniques \cite{bereska2024mechanistic} are emerging as powerful tools to reverse-engineer the computational mechanisms of neural networks into human-understandable concepts. However, scaling these techniques to MLLMs and applying them to complex medical tasks remains an open challenge. 
\begin{figure}[h]
    \centering
    \includegraphics[width=\linewidth]{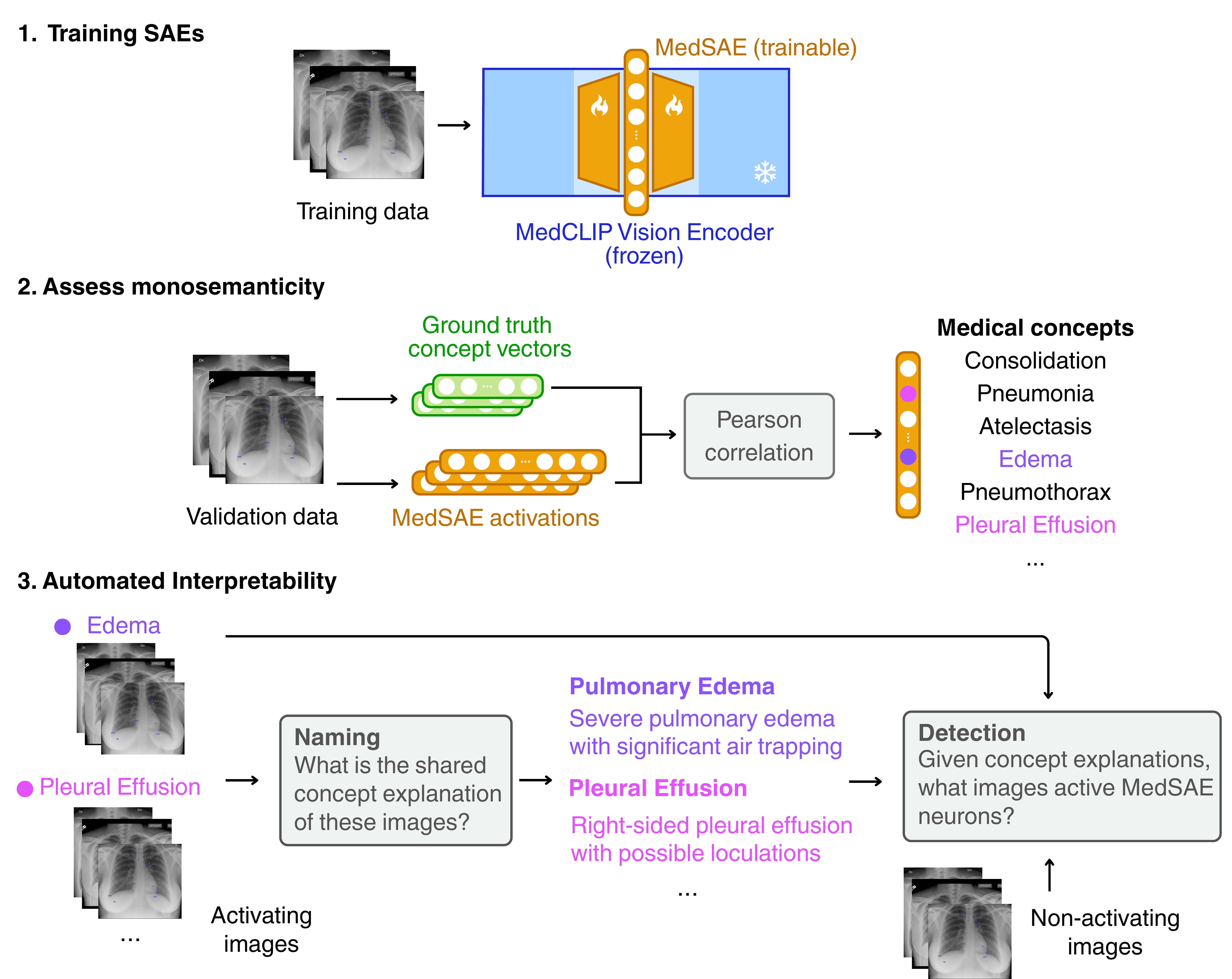}
    \caption{
The overall proposed pipeline. (1) We first train MedSAE from MedCLIP vision encoder and extract corresponding embeddings. (2) Then, we compute their Pearson correlation with multi-hot encoded vector labels to identify MedSAE neurons-concepts mappings. (3) As a final step, for each MedSAE neuron, top-activating images are used to generate concept names via structured prompting. These names are validated through a detection task, where MedGemma yields a quantitative measure of semantic alignment.
}
    \label{fig:full_overview}
\end{figure}
In this work, as shown in Figure~\ref{fig:full_overview}, we present the first empirical analysis of mechanistic interpretability in medical MLLMs using Medical Sparse Autoencoders (MedSAEs) to uncover meaningful internal representations from MedCLIP~\cite{wang2022medclip}, trained on chest X-rays. We propose an automated feature naming via MedGemma~\cite{medgemma2025}, a foundation vision-language model that assigns and validates human-interpretable labels to individual MedSAE neurons. 

Our experiments show that MedSAE features exhibit higher class selectivity and stronger alignment with clinical concepts than those from raw MedCLIP embeddings. Notably, we identify and name 21 medically meaningful features using MedGemma with high interpretability accuracy. 

To summarize, our main contributions are as follows:
\begin{itemize}
    \item We propose a novel evaluation framework that combines correlation-based metrics (Section~\ref{sec:sae-mono}) with automated concept naming and validation using the medical foundation model MedGemma, for a quantitative assessment of interpretability (Sections~\ref{sec:sae-mono} and \ref{sec:method:medgemma}).
    \item We show that MedCLIP embeddings approximately satisfy the linear superposition hypothesis through an averaging-based probing approach (Section~\ref{sec:linearity_sanity_check}).
    \item We show that SAEs can extract medically meaningful and interpretable features from the internal representations of a medical vision-language model (Section~\ref{sec:results:correlation}).
\end{itemize}
\section{Background and related work}
\label{sec:background}

Mechanistic interpretability aims to decompose deep learning models into human-understandable components by analyzing their model activations. A central assumption underlying many recent advances is the \textit{linear representation hypothesis}~\cite{park2023linear}, which posits that high-level features are represented as approximately linear directions in activation space, enabling meaningful interventions through linear operations. However, empirical evidence suggests that neural networks operate in a regime of \textit{superposition}~\cite{bereska2024mechanistic}, where the number of learned features exceeds the dimensionality of the representation space, forcing multiple features to be multiplexed across shared neurons. In this setting, individual neurons are no longer aligned with single, interpretable concepts, complicating attribution and targeted intervention. 

\subsection{Sparse Autoencoders}\label{sec:sae}
Sparse autoencoders (SAEs) have recently emerged as a practical tool to address superposition by learning overcomplete, sparse representations that disentangle superposed features into a larger set of latent directions. By mapping dense activations to sparse, approximately monosemantic latent units, SAEs provide a concrete mechanism for recovering linear feature directions consistent with the linear representation hypothesis, while offering a tractable interface for mechanistic analysis and causal manipulation of neural networks.

A single-layer ReLU SAE operates on $d$-dimensional activation vectors. Let $\mathbf{x} \in \mathbb{R}^d$ denote the input activation vector and $m$ be the SAE latent dimension, typically set to $d$ multiplied by a positive expansion factor. The encoder and decoder are defined as:
\begin{equation}
\begin{split}
\mathbf{z} &= \text{ReLU}(\mathbf{W}_{enc}(\mathbf{x}-\mathbf{b}_{pre}) + \mathbf{b}_{enc}) \\
\hat{\mathbf{x}} &= \mathbf{W}_{dec}\mathbf{z} + \mathbf{b}_{pre},
\end{split}
\end{equation}
where $\mathbf{W}_{enc} \in \mathbb{R}^{m \times d}$ and $\mathbf{W}_{dec} \in \mathbb{R}^{d \times m}$ are the encoder and decoder weight matrices respectively, and $\mathbf{b}_{pre} \in \mathbb{R}^d$ and $\mathbf{b}_{enc} \in \mathbb{R}^m$ are learnable bias terms.

To encourage sparsity in the latent representation $\mathbf{z}$, an L1 penalty is added to the objective:
\begin{equation}
\mathcal{L} := \|\mathbf{x} - \mathbf{\hat{x}}\|_2^2 + \lambda \|\mathbf{z}\|_1,
\end{equation}
where $\lambda > 0$ controls regularization strength.
Our work demonstrates the potential of SAEs by training a ReLU-based model on chest X-ray language-vision embeddings to disentangle superposed features and unveil meaningful medical concepts within medical language-vision models.

\subsection{Neurons-concept matching}
Although SAEs are effective in learning monosemantic representations, interpreting these neurons—i.e., mapping them to human-understandable concepts—remains non-trivial. Previous work in language models used LLMs for identifying neuron-encoded concepts~\cite{bills2023language} and was extended to vision models by leveraging recent Vision-
Language Models (VLMs) to summarize a set of images and captions and find their shared concept~\cite{StanfordHAI2025}. More efficient methods~\cite{rao2024discover, zaigrajew2025} were also proposed for CLIP-trained SAE, which leverages CLIP's representation space. These methods use a predefined vocabulary of concepts (e.g., `hair', `pink') to compute cosine similarity between CLIP embeddings and SAE decoder columns. For the feature columns in the SAE decoder,the best matching concept to the neuron is determined by maximizing cosine similarity, where a value of 1 indicates perfect alignment. 

Due to the difficulty of constructing a robust medical vocabulary for decoder-based concept matching, we instead propose a new evaluation framework (Section~\ref{sec:method:medgemma}) that uses MedGemma to identify and summarize neuron-activating image concepts.

\subsection{Mechanistic Interpretability for Healthcare}
While the literature on mechanistic interpretability in healthcare remains relatively limited, there is a growing interest in applying SAEs to reveal interpretable representations in medical imaging models. For instance, SAEs are employed to debug melanoma detection networks, uncovering dataset biases through neuron activation patterns~\cite{dreyer2025semanticlens}. Similarly, SAEs are also used to generate radiology reports by mapping image tokens into interpretable features, achieving both transparency and competitive diagnostic performance~\cite{abdulaal2024xrayworth15features}. More recently,  SAEs are applied to vision transformers in histopathology, identifying biologically meaningful features, such as specific cell types, with improved robustness to confounding factors~\cite{le2024learningbiologicallyrelevantfeatures}. A major challenge across these works lies in naming and semantically grounding the discovered features. Building on this foundation, our work further investigates SAEs in medical vision models, specifically for radiological imaging, by training a ReLU-based SAE on MedCLIP, a vision-language model for chest X-rays, and introducing a domain-specific framework for neuron naming and interpretability evaluation.

\section{Methodology}
\label{sec:method}
We now present our methodology for extracting and evaluating interpretable latent representations using SAEs. As illustrated in Figure~\ref{fig:full_overview}, our MedSAE pipeline comprises three main stages: (1) training SAEs on MedCLIP embeddings, (2) assessing neuron monosemanticity, and (3) performing automated interpretability and neuron naming. Each step is detailed in the following subsections.

\subsection{Image Embeddings Extraction and Training}
Image embeddings are computed using the MedCLIP image backbone. MedCLIP embeddings can exhibit misalignment across modalities, which can impact SAE training convergence and cross-modal transferability. 
Following \cite{bhalla2024interpreting}, we normalize embeddings to ensure consistent behavior across modalities. 
We first center embeddings by subtracting the per-modality mean estimated from the training dataset. 
Next, we scale the centered embeddings by a dataset-computed scaling factor to obtain $\mathbb{E}_{\mathbf{x} \in \mathcal{X}} [\|\mathbf{x}\|_2] = \sqrt{d}$ with $d$ the dimension of the MedCLIP embeddings. 
We then train ReLU-based SAEs on these normalized embeddings, encouraging sparse and disentangled activations. This scaling ensures that $\lambda$ has consistent effects across different CLIP architectures and modalities. 

\subsection{Assessing SAEs Monosemanticity}\label{sec:sae-mono}
As a first step toward validating monosemanticity and the alignment of neurons with clinically meaningful concepts, we leverage label-based correlation analyses to identify neurons that are selectively responsive to specific medical findings.  Inspired by the approach in \cite{le2024learningbiologicallyrelevantfeatures}, we use the Pearson correlation coefficient to quantify the strength of the linear relationship between each neuron's activation and the presence or absence of a particular medical finding. Let $\mathbf{Z} \in \mathbb{R}^{n \times m}$ be the activation matrix of the SAE, where $n$ is the number of samples and $m$ the number of neurons. Let $\mathbf{Y} \in \mathbb{R}^{n \times k}$ be the corresponding multi-hot encoded label matrix, with $k$ the number of labels. The Pearson correlation between neuron $i$ and label $j$ is computed as:

\begin{equation}
    \rho_{i,j} = \frac{\text{Cov}(\mathbf{Z}_i, \mathbf{Y}_j)}{\sigma_{\mathbf{Z}_i} \sigma_{\mathbf{Y}_j}}
\end{equation}

To evaluate whether a neuron is \textit{monosemantic} (i.e., predominantly associated with a single concept) we compute the entropy of its correlation distribution across all labels. Specifically, we define:
\begin{equation}
p_{i,j} = \frac{|\rho_{i,j}|}{\sum_k |\rho_{i,k}|}
\end{equation}
Then, the entropy of neuron $i$ is given by:
\begin{equation}
H_i = -\sum_j p_{i,j} \log_2 p_{i,j}
\end{equation}

A lower entropy $H_i$ indicates that neuron $i$ is sharply peaked on a single concept, suggesting high monosemanticity. In contrast, higher entropy implies more distributed activation across multiple concepts. In practice, we use the average entropy across MedSAE neurons as a selection criterion to compare different configurations and identify the model that achieves the highest degree of monosemanticity.

\subsection{Automated Naming with MedGemma}\label{sec:method:medgemma}
We leverage the capabilities of MedGemma to generate semantic interpretations for groups of SAE activations. Given a set of X-ray images that strongly activate a particular latent feature, we prompt MedGemma to identify and summarize the shared concept underlying these samples. These interpretations can reflect medical phenomena (e.g., a specific pathology or anatomical abnormality) or non-medical characteristics (e.g., image artifacts, patient positioning, or device presence). This automated naming framework enables us to attach human-readable labels to otherwise abstract SAE neurons, serving as a first step toward interpretable latent representations in the medical domain.

While automated naming provides semantic anchors for SAE features, it does not inherently guarantee accuracy or clinical relevance. Therefore, we incorporate validation strategies to assess the consistency and interpretability of the generated names. To this end, we adopt the detection metric introduced by \cite{paulo2024autoname}. In this setup, MedGemma is provided with a generated interpretation and a balanced set of activating and non-activating images. It is then asked to identify which images match the concept described—effectively performing a binary classification task. This allows us to assess whether the interpretation genuinely reflects a consistent and isolatable feature in the image space. 

\section{Experiments and Results}
\label{sec:results}
This section evaluates our previously introduced method, showing that MedSAEs effectively disentangle superposed representations in MedCLIP embeddings, revealing structured and clinically meaningful features. These results support the potential of SAEs as a practical tool for mechanistic interpretability in vision-language models, bridging the gap between deep feature representations and human-understandable medical concepts. We provide the source code in the supplementary and will release it publicly upon acceptance.

\subsection{Dataset and implementation details}
To ensure consistency with the MedCLIP data distribution, our study employed CheXpert dataset~\cite{irvin2019chexpert}. We curated a balanced evaluation dataset from CheXpert, comprising exactly 200 images per class. We trained our ReLU-MedSAEs on the last layer of a MedCLIP-ResNet ($d=512$) on the original CheXpert train split for 200 epochs. We choose a learning rate of $7\cdot10^{-6}$, a L1 penalty coefficient $\lambda$ of $3\cdot10^{-4}$ and an expansion rate of $16$ based on a hyperparameter search over the SAE loss. 



\subsection{Sanity checking linearity of MedCLIP Embeddings.}\label{sec:linearity_sanity_check}
Applying SAEs to uncover interpretable concepts in MedCLIP presupposes the validity of the linear representation and superposition hypotheses, assumptions that are not trivial in the medical imaging domain. Inspired by~\cite{bhalla2024interpreting}, we examined whether the embedding of a composite input approximates the average of its individual components. To do this, we constructed composite images by placing two input images, $\mathbf{x}_a$ and $\mathbf{x}_b$, in opposing quadrants of a blank canvas to form $\mathbf{x}_{ab}$. We then embedded $\mathbf{x}_a$, $\mathbf{x}_b$, and $\mathbf{x}_{ab}$ using MedCLIP, yielding representations $\mathbf{v}_a$, $\mathbf{v}_b$, and $\mathbf{v}_{ab}$. By solving for scalar weights $w_a$ and $w_b$ in the equation $w_a \mathbf{v}_a + w_b \mathbf{v}_b = \mathbf{v}_{ab}$, we assessed the degree to which $\mathbf{v}_{ab}$ could be linearly reconstructed from its components. Table~\ref{tab:clip_lin} reports the learned weights and the cosine similarity between the predicted composite embedding $\hat{\mathbf{v}}_{ab} = [\mathbf{v}_a, \mathbf{v}_b] \cdot [w_a, w_b]$ and the actual $\mathbf{v}_{ab}$. 

\begin{table}[h]
\caption{Sanity checking the linearity of MedCLIP Embeddings.}
\label{tab:clip_lin}
\begin{center}
\begin{small}
\begin{sc}
\begin{tabular}{lcccr}
\toprule
           & $w_a$     & $w_b$     & cosine($\hat{\mathbf{v}}, \mathbf{v}$) \\ 
\midrule
CheXpert   & 0.49 $\pm$ 0.32 & 0.49 $\pm$ 0.31 & 0.97 $\pm$ 0.022       \\
\bottomrule
\end{tabular}
\end{sc}
\end{small}
\end{center}
\end{table}

Using 100{,}000 randomly sampled image pairs from the 200k CheXpert training set, we find that the learned weights $w_a$ and $w_b$ are roughly equal and centered around 0.5, with notable variability ($\text{std} \approx 0.32$). Despite this, the high cosine similarity ($0.97 \pm 0.022$) between predicted and actual composite embeddings supports the linear superposition hypothesis in MedCLIP’s embedding space.

\subsection{Interpretable Feature Disentanglement: Monosemanticity of SAE Neurons and Medical Class Correlations}
\label{sec:results:correlation}

\begin{figure}[h]
    \centering
    \includegraphics[width=1\linewidth]{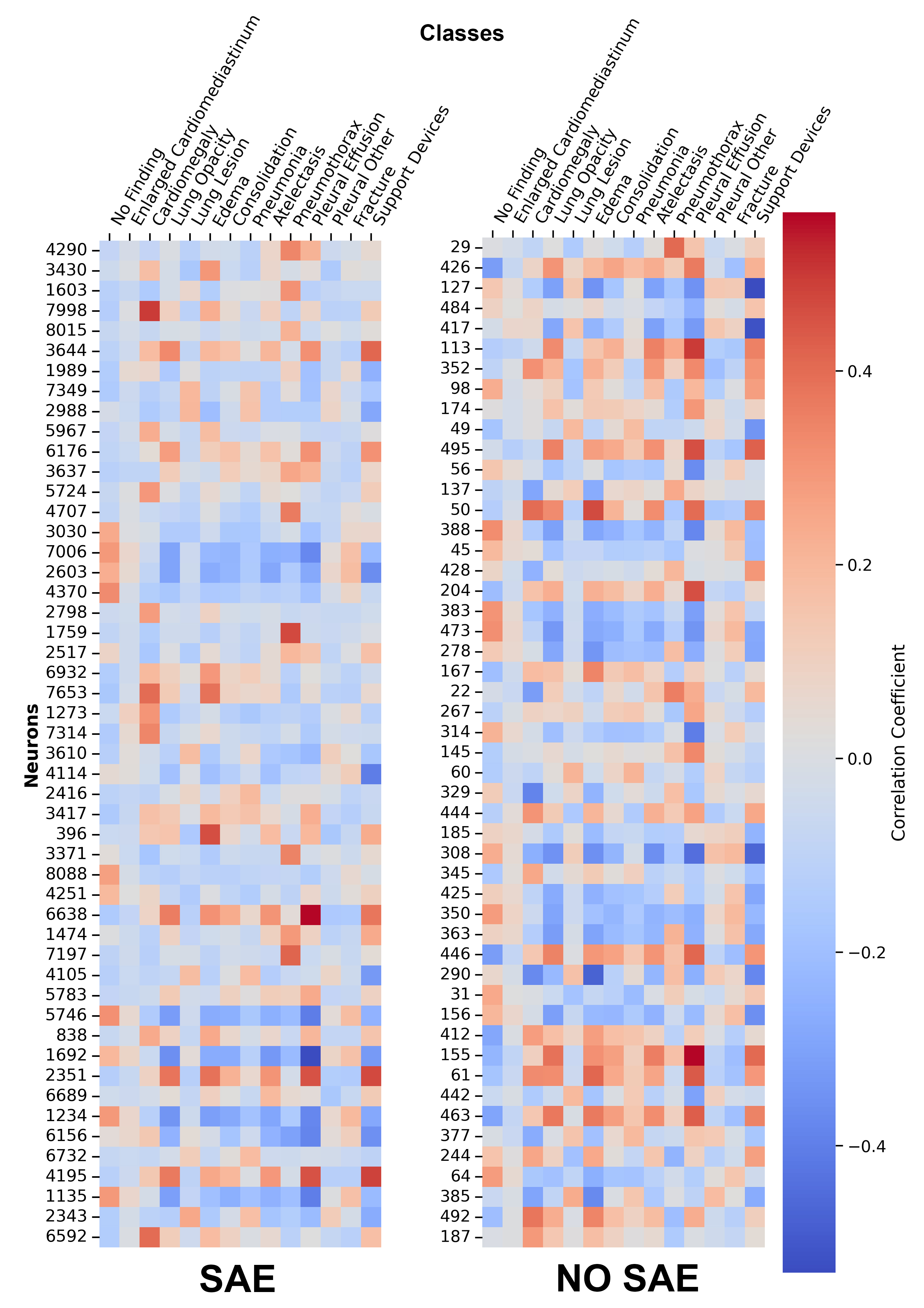}
    \caption{Heatmap of Pearson correlation coefficients between 100 random neurons and the corresponding medical classes in the CheXpert14x200 dataset. SAE neurons exhibit more concentrated correlations with individual classes compared to raw MedCLIP embeddings.}
    \label{fig:neuron_class_correlation}
\end{figure}

We select the model that achieves the best neuron-class entropy on CheXpert14x200. This model also maintains a favorable sparsity–reconstruction balance, with a 0.20\% L0 activation rate and a 0.98 fraction of variance explained (FVE) despite a significant amount of dead neurons (30\%). Our results
align with reasonable sparsity settings as identified by~\cite{gao2024scaling}. However, these metrics do not directly assess whether the resulting representations are interpretable or monosemantic. We analyze how the sparsity–reconstruction trade-off relates to interpretability metrics in Appendix 7.1. As expected, we find that neurons from the SAE exhibit lower entropy compared to MedCLIP’s raw embeddings, indicating improved monosemantic behavior (2.25 vs 2.38). This trend is also visually depicted in Figure~\ref{fig:neuron_class_correlation}.

\subsection{Discovering Medical Concepts via MedGemma-Based Automated Interpretability}
\label{sec:results:medgemma}
Our previous analyses suggest that some SAE neurons exhibit strong correlations with specific medical concepts, indicative of monosemanticity. However, while such patterns point to interpretability, they do not reveal the real semantics of each neuron. To uncover these latent concepts, we apply the MedGemma-Based Automated Interpretability framework introduced in Section~\ref{sec:method:medgemma}. Appendix 7.2 reports the prompts used in the MedGemma.

\begin{figure}[h]
    \centering
    \includegraphics[width=1\linewidth]{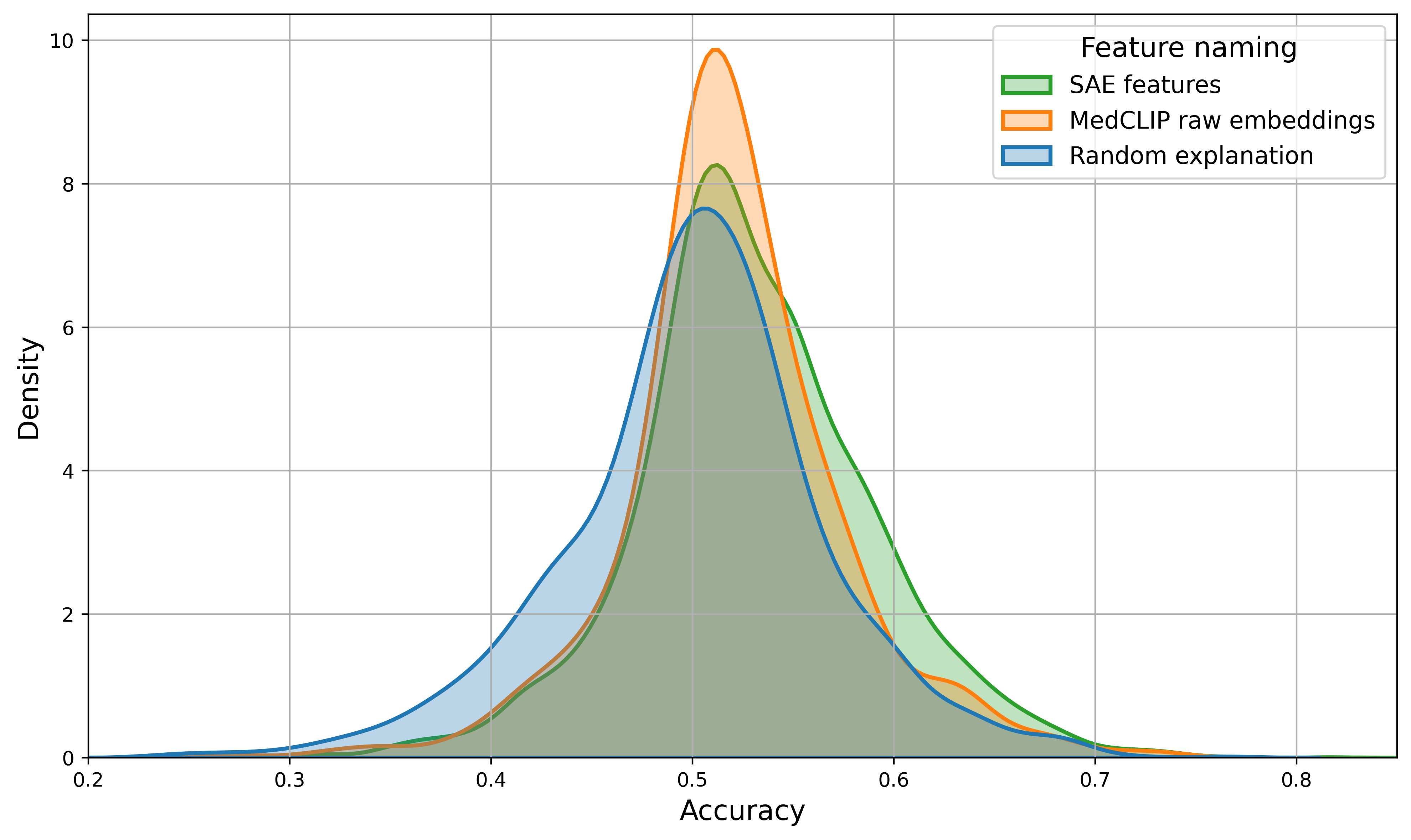}
    \caption{
    Density of interpretation detection accuracy for three embedding types: SAE neurons, raw MedCLIP embeddings, and SAE with shuffled neuron-concept pairs. SAE features yield a higher density of neurons with accurate and consistent interpretations, indicating stronger semantic alignment. The performance drop in the shuffled case confirms that meaningful interpretations do not arise randomly.
    }
    \label{fig:naming_eval_accuracy}
\end{figure}

We conducted this evaluation across three conditions: (1) neurons from our trained SAE model, (2) raw MedCLIP embeddings, and (3) a control condition where neuron-concept pairs from the SAE were randomly shuffled to test for statistical artifacts. The density–accuracy curves of this detection task are presented in Figure~\ref{fig:naming_eval_accuracy}. While the overall shape of the distributions is similar, SAE neurons yield a noticeably higher density at high detection accuracies compared to both MedCLIP and the shuffled control. This suggests that SAE neurons are not only more interpretable but also more consistently associated with distinct medical concepts. Importantly, the drop in performance for the shuffled case confirms that meaningful interpretations are not emerging by chance.

To further analyze interpretability at a finer scale, we examined the individual neurons with the highest interpretation accuracy. As shown in Table~\ref{tab:top_named_neurons_unique}, we identified 21 different SAE neuron interpretations with detection accuracy above 70\%, in contrast to just two for MedCLIP embeddings. Many of these neurons align with clinically coherent concepts such as \textit{severe pulmonary edema}, \textit{right-sided pleural effusion}, or \textit{cardiomegaly with vascular congestion}. 
Overall, these findings provide additional evidence that the SAE architecture promotes disentanglement, giving rise to individual neurons that robustly encode specific and medically meaningful concepts.

\begin{table}[h!]
\caption{Top SAE neurons and their corresponding MedGemma-generated concepts, sorted by detection accuracy. Duplicate concept descriptions have been removed.}
\centering
\resizebox{\linewidth}{!}{
\begin{tabular}{lc|p{10cm}}
\toprule
\textbf{Neuron} & \textbf{Accuracy} & \textbf{Concept Description} \\
\midrule
1215 & 0.82 & Severe pulmonary edema with significant air trapping and subcutaneous emphysema \\
7468 & 0.77 & Right upper lobe consolidation with possible cavitation \\
2320 & 0.75 & Right-sided pleural effusion with possible loculations \\
876  & 0.75 & Diffuse bilateral infiltrates with significant opacification of the lung fields, likely due to pulmonary edema or infectious process \\
85   & 0.73 & Diffuse bilateral interstitial lung disease (ILD) with multiple lines of central venous access \\
3826 & 0.73 & Pulmonary edema with cardiomegaly \\
4214 & 0.73 & Severe cardiomegaly with significant pulmonary vascular congestion \\
283  & 0.73 & Severe right-sided pneumothorax with significant subcutaneous emphysema and multiple lines/tubes in place \\
817  & 0.73 & Right-sided pleural effusion with loculations \\
1859 & 0.73 & Large left pleural effusion with possible loculation \\
5929 & 0.73 & Diffuse bilateral pulmonary infiltrates \\
127  & 0.72 & Pulmonary edema with significant bilateral opacities \\
5616 & 0.72 & Normal chest X-ray with clear visualization of lungs, heart, and rib cage in upright position \\
1251 & 0.72 & Cardiac device placement and associated pulmonary edema \\
2378 & 0.72 & Severe bilateral pulmonary infiltrates with significant opacification of both lung fields, likely due to edema or infection \\
2862 & 0.72 & Pulmonary edema with significant pulmonary vascular congestion and cardiomegaly \\
8153 & 0.72 & Right-sided pleural effusion with significant underlying lung opacity \\
4728 & 0.70 & Cardiomegaly with pulmonary edema \\
1266 & 0.70 & Diffuse bilateral pulmonary infiltrates with central line placement \\
2708 & 0.70 & Right-sided pleural effusion with pulmonary edema \\
6552 & 0.70 & Normal adult chest radiograph \\
\bottomrule
\end{tabular}
}
\label{tab:top_named_neurons_unique}
\end{table}

\section{Conclusion}
\label{sec:conclusion}

This work takes a first step toward mechanistic interpretability in medical vision-language models by leveraging sparse autoencoders to extract clinically meaningful and approximately monosemantic features from MedCLIP embeddings. We introduce an evaluation framework that combines correlation-based metrics with automated neuron naming via MedGemma, demonstrating improved interpretability and alignment with clinical concepts on the CheXpert dataset.

Despite these promising results, several limitations remain. We adopt a standard ReLU-based SAE, which entails a trade-off between sparsity, overcompleteness, and feature utilization; inactive units may reflect this capacity--sparsity tension under superposition rather than a failure of the approach. Recent work suggests that classical dictionary-learning objectives can be suboptimal~\cite{oneill2025suboptimal}, motivating future exploration of alternative architectures such as BatchTopK and Matryoshka SAEs. Moreover, our analysis relies on the linear representation and superposition hypotheses, which may not universally hold, and extending interpretability analyses across multiple layers could reveal nonlinear feature interactions and circuits. Our automated neuron naming procedure, while scalable, is computationally intensive and sensitive to the selection of top-activating samples, potentially introducing dataset-specific biases. Human-in-the-loop validation, particularly involving medical experts, remains necessary to assess clinical relevance and safety. In addition, our experiments are limited to the MedCLIP-ResNet backbone, and the robustness and generality of learned features across architectures, modalities, institutions, and patient populations remain open questions.

Overall, our findings highlight sparse autoencoders as a promising tool for uncovering interpretable structure in medical foundation models, and we hope this work motivates further research toward more transparent, robust, and trustworthy medical AI systems.

\bibliographystyle{IEEEbib}
\bibliography{bib}

@String(ECCV= {Eur. Conf. Comput. Vis.})

@String(AAAI = {AAAI})

@String(ECCV  = {ECCV})

@article{LITJENS201760,
    title = {A survey on deep learning in medical image analysis},
    journal = {Medical Image Analysis},
    volume = {42},
    pages = {60-88},
    year = {2017},
    issn = {1361-8415},
    doi = {https://doi.org/10.1016/j.media.2017.07.005},
    url = {https://www.sciencedirect.com/science/article/pii/S1361841517301135},
    author = {Geert Litjens and Thijs Kooi and Babak Ehteshami Bejnordi and Arnaud Arindra Adiyoso Setio and Francesco Ciompi and Mohsen Ghafoorian and Jeroen A.W.M. {van der Laak} and Bram {van Ginneken} and Clara I. Sánchez}
}

@article{bereska2024mechanistic,
    title={Mechanistic Interpretability for {AI} Safety - A Review},
    author={Leonard Bereska and Stratis Gavves},
    journal={Transactions on Machine Learning Research},
    issn={2835-8856},
    year={2024},
    url={https://openreview.net/forum?id=ePUVetPKu6},
    note={Survey Certification, Expert Certification}
}

@inproceedings{le2024learningbiologicallyrelevantfeatures,
    title={Learning biologically relevant features in a pathology foundation model using sparse autoencoders},
    author={Nhat Minh Le and Neel Patel and Ciyue Shen and Blake Martin and Alfred Eng and Chintan Shah and Sean Grullon and Dinkar Juyal},
    booktitle={Advancements In Medical Foundation Models: Explainability, Robustness, Security, and Beyond},
    year={2024},
    url={https://openreview.net/forum?id=daV16mhUBd}
}

@inproceedings{paulo2024autoname,
    title={Automatically Interpreting Millions of Features in Large Language Models},
    author={Gon{\c{c}}alo Santos Paulo and Alex Troy Mallen and Caden Juang and Nora Belrose},
    booktitle={Forty-second International Conference on Machine Learning},
    year={2025},
    url={https://openreview.net/forum?id=EemtbhJOXc}
}

@inproceedings{bhalla2024interpreting,
    title={Interpreting {CLIP} with Sparse Linear Concept Embeddings (SpLi{CE})},
    author={Usha Bhalla and Alex Oesterling and Suraj Srinivas and Flavio Calmon and Himabindu Lakkaraju},
    booktitle={The Thirty-eighth Annual Conference on Neural Information Processing Systems},
    year={2024},
    url={https://openreview.net/forum?id=7UyBKTFrtd}
}

@misc{medgemma2025,
    title={MedGemma Technical Report}, 
    author={Andrew Sellergren and Sahar Kazemzadeh and Tiam Jaroensri and Atilla Kiraly and Madeleine Traverse and others},
    year={2025},
    eprint={2507.05201},
    archivePrefix={arXiv},
    primaryClass={cs.AI},
    url={https://arxiv.org/abs/2507.05201}, 
}

@article{StanfordHAI2025,
  author = {Stanford HAI},
  title = {Finding Monosemantic Subspaces and Human-Compatible Interpretations in Vision Transformers through Sparse Coding},
  journal = {Stanford HAI},
  year = {2025},
  url = {https://hai.stanford.edu/research/finding-monosemantic-subspaces-and-human-compatible-interpretations-in-vision-transformers-through-sparse-coding},
  note = {Accessed on: 2025-01-01}
}

@misc{oneill2025suboptimal,
    title={Compute Optimal Inference and Provable Amortisation Gap in Sparse Autoencoders},
    author={Charles O'Neill and David Klindt},
    year={2024},
    url={https://openreview.net/forum?id=ghH6YYDs15}
}

@inproceedings{zaigrajew2025,
    title={Interpreting {CLIP} with Hierarchical Sparse Autoencoders},
    author={Vladimir Zaigrajew and Hubert Baniecki and Przemyslaw Biecek},
    booktitle={Forty-second International Conference on Machine Learning},
    year={2025},
    url={https://openreview.net/forum?id=5MQQsenQBm}
}

@InProceedings{rao2024discover,
    author="Rao, Sukrut
    and Mahajan, Sweta
    and B{\"o}hle, Moritz
    and Schiele, Bernt",
    editor="Leonardis, Ale{\v{s}}
    and Ricci, Elisa
    and Roth, Stefan
    and Russakovsky, Olga
    and Sattler, Torsten
    and Varol, G{\"u}l",
    title="Discover-then-Name: Task-Agnostic Concept Bottlenecks via Automated Concept Discovery",
    booktitle="Computer Vision -- ECCV 2024",
    year="2024",
    publisher="Springer Nature Switzerland",
    address="Cham",
    pages="444--461",
    isbn="978-3-031-72980-5"
}

@misc{bills2023language,
    title={Language models can explain neurons in language models},
    author={
    Bills, Steven and Cammarata, Nick and Mossing, Dan and Tillman, Henk and Gao, Leo and Goh, Gabriel and Sutskever, Ilya and Leike, Jan and Wu, Jeff and Saunders, William
    },
    year={2023},
    month={May},
    howpublished = {\url{https://openaipublic.blob.core.windows.net/neuron-explainer/paper/index.html}}
}

@article{irvin2019chexpert, 
    author = {Irvin, Jeremy and Rajpurkar, Pranav and Ko, Michael and Yu, Yifan and Ciurea-Ilcus, Silviana and Chute, Chris and Marklund, Henrik and Haghgoo, Behzad and Ball, Robyn and Shpanskaya, Katie and Seekins, Jayne and Mong, David A. and Halabi, Safwan S. and Sandberg, Jesse K. and Jones, Ricky and Larson, David B. and Langlotz, Curtis P. and Patel, Bhavik N. and Lungren, Matthew P. and Ng, Andrew Y.}, 
    title={CheXpert: A Large Chest Radiograph Dataset with Uncertainty Labels and Expert Comparison}, 
    volume={33}, 
    url={https://ojs.aaai.org/index.php/AAAI/article/view/3834}, DOI={10.1609/aaai.v33i01.3301590}, 
    number={01}, 
    journal={Proceedings of the AAAI Conference on Artificial Intelligence}, year={2019}, 
    month={Jul.}, pages={590-597}  
}

@misc{abdulaal2024xrayworth15features,
      title={An X-Ray Is Worth 15 Features: Sparse Autoencoders for Interpretable Radiology Report Generation}, 
      author={Ahmed Abdulaal and Hugo Fry and Nina Montaña-Brown and Ayodeji Ijishakin and Jack Gao and Stephanie Hyland and Daniel C. Alexander and Daniel C. Castro},
      year={2024},
      eprint={2410.03334},
      archivePrefix={arXiv},
      primaryClass={cs.CV},
      url={https://arxiv.org/abs/2410.03334}, 
}

@article{dreyer2025semanticlens,
    title = {Mechanistic understanding and validation of large {AI} models with {SemanticLens}},
    copyright = {2025 The Author(s)},
    issn = {2522-5839},
    url = {https://www.nature.com/articles/s42256-025-01084-w},
    doi = {10.1038/s42256-025-01084-w},
    language = {en},
    urldate = {2025-08-18},
    journal = {Nature Machine Intelligence},
    author = {Dreyer, Maximilian and Berend, Jim and Labarta, Tobias and Vielhaben, Johanna and Wiegand, Thomas and Lapuschkin, Sebastian and Samek, Wojciech},
    month = aug,
    year = {2025},
    note = {Publisher: Nature Publishing Group},
    pages = {1--14}, 
}

@inproceedings{wang2022medclip,
    title = "{M}ed{CLIP}: Contrastive Learning from Unpaired Medical Images and Text",
    author = "Wang, Zifeng  and
      Wu, Zhenbang  and
      Agarwal, Dinesh  and
      Sun, Jimeng",
    editor = "Goldberg, Yoav  and
      Kozareva, Zornitsa  and
      Zhang, Yue",
    booktitle = "Proceedings of the 2022 Conference on Empirical Methods in Natural Language Processing",
    month = dec,
    year = "2022",
    address = "Abu Dhabi, United Arab Emirates",
    publisher = "Association for Computational Linguistics",
    url = "https://aclanthology.org/2022.emnlp-main.256/",
    doi = "10.18653/v1/2022.emnlp-main.256",
    pages = "3876--3887"
}

@inproceedings{gao2024scaling,
    title={Scaling and evaluating sparse autoencoders},
    author={Leo Gao and Tom Dupre la Tour and Henk Tillman and Gabriel Goh and Rajan Troll and Alec Radford and Ilya Sutskever and Jan Leike and Jeffrey Wu},
    booktitle={The Thirteenth International Conference on Learning Representations},
    year={2025},
    url={https://openreview.net/forum?id=tcsZt9ZNKD}
}

@inproceedings{park2023linear,
    title={The Linear Representation Hypothesis and the Geometry of Large Language Models},
    author={Kiho Park and Yo Joong Choe and Victor Veitch},
    booktitle={Forty-first International Conference on Machine Learning},
    year={2024},
    url={https://openreview.net/forum?id=UGpGkLzwpP}
}

\clearpage
\section{Appendix}
\subsection{Linking Traditional SAE Metrics with Interpretability and Monosemanticity}
\label{appendix:sae_metrics_interpr}
\begin{figure}[h!]
    \centering
    \includegraphics[width=1\linewidth]{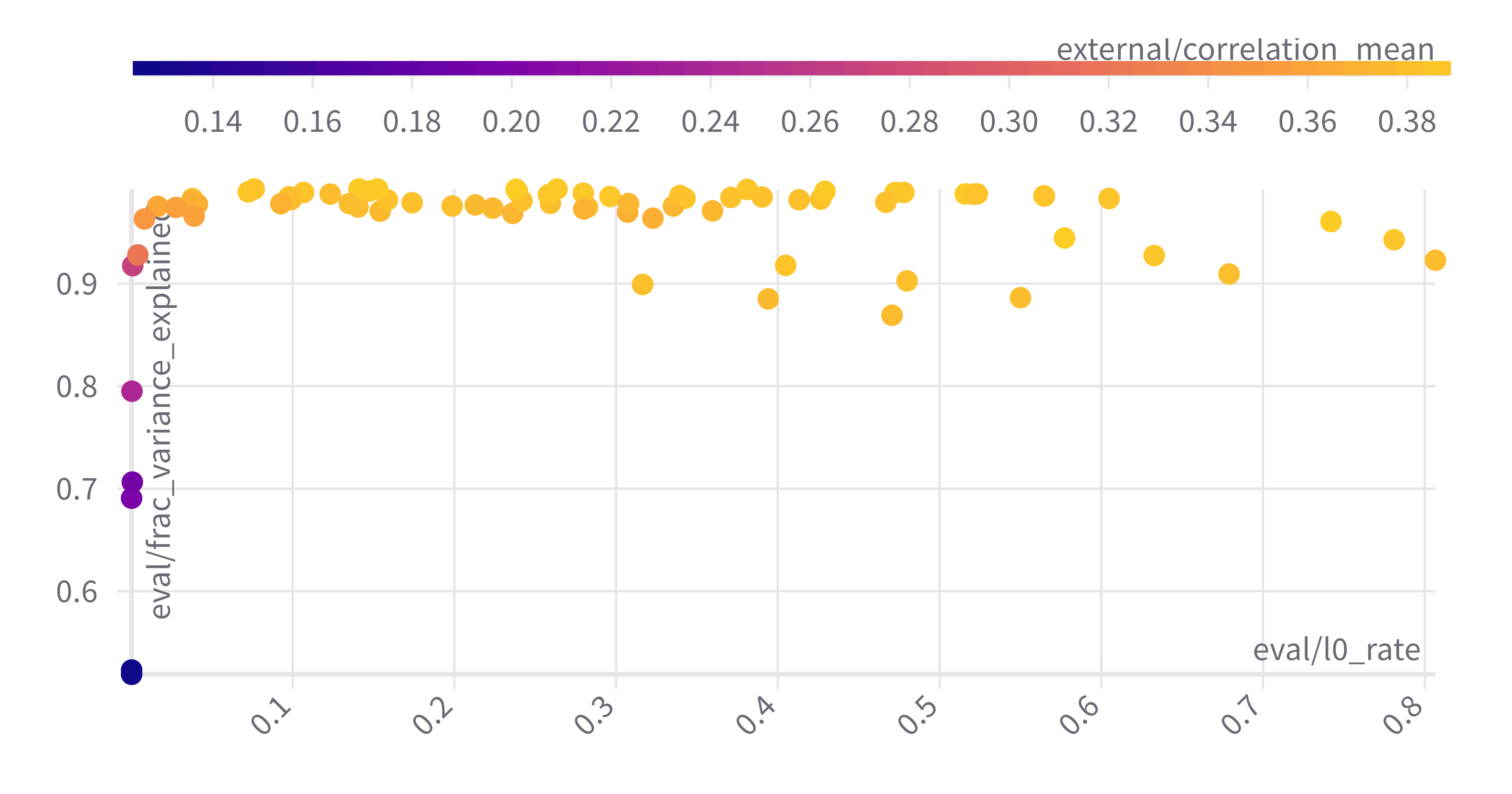}
    \caption{Relationship between the sparsity–reconstruction trade-off and mean top-class correlation.}
    \label{fig:l0-rec-corr}
\end{figure}

\begin{figure}[h!]
    \centering
    \includegraphics[width=1\linewidth]{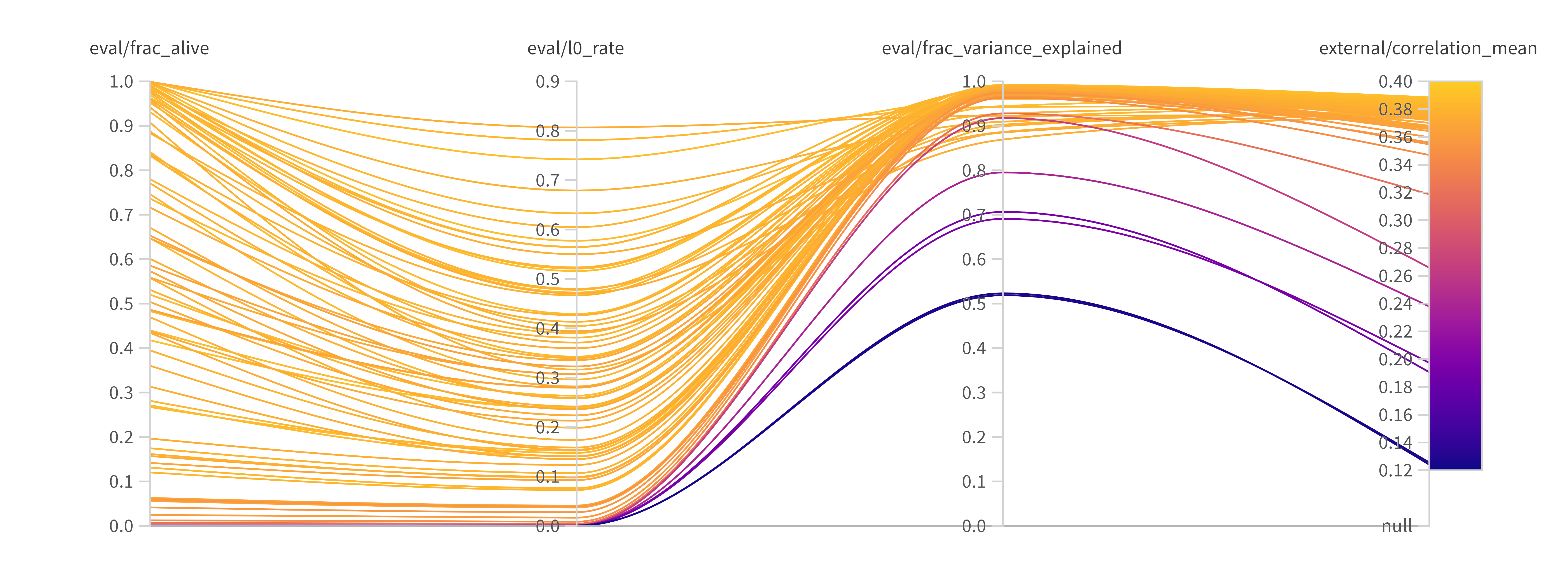}
    \caption{Mean correlation with respect to sparsity, reconstruction quality, and proportion of active neurons.}
    \label{fig:l0-rec-corr-frac}
\end{figure}

Here we analyze how the sparsity–reconstruction trade-off relates to interpretability metrics—specifically, average class correlation and correlation entropy. Figures~\ref{fig:l0-rec-corr} and~\ref{fig:l0-rec-corr-frac} show that models with better reconstruction performance tend to exhibit higher mean top-class correlation, suggesting that more informative reconstructions may lead to features that align more closely with class-specific signals.

\begin{figure}[h!]
    \centering
    \includegraphics[width=1\linewidth]{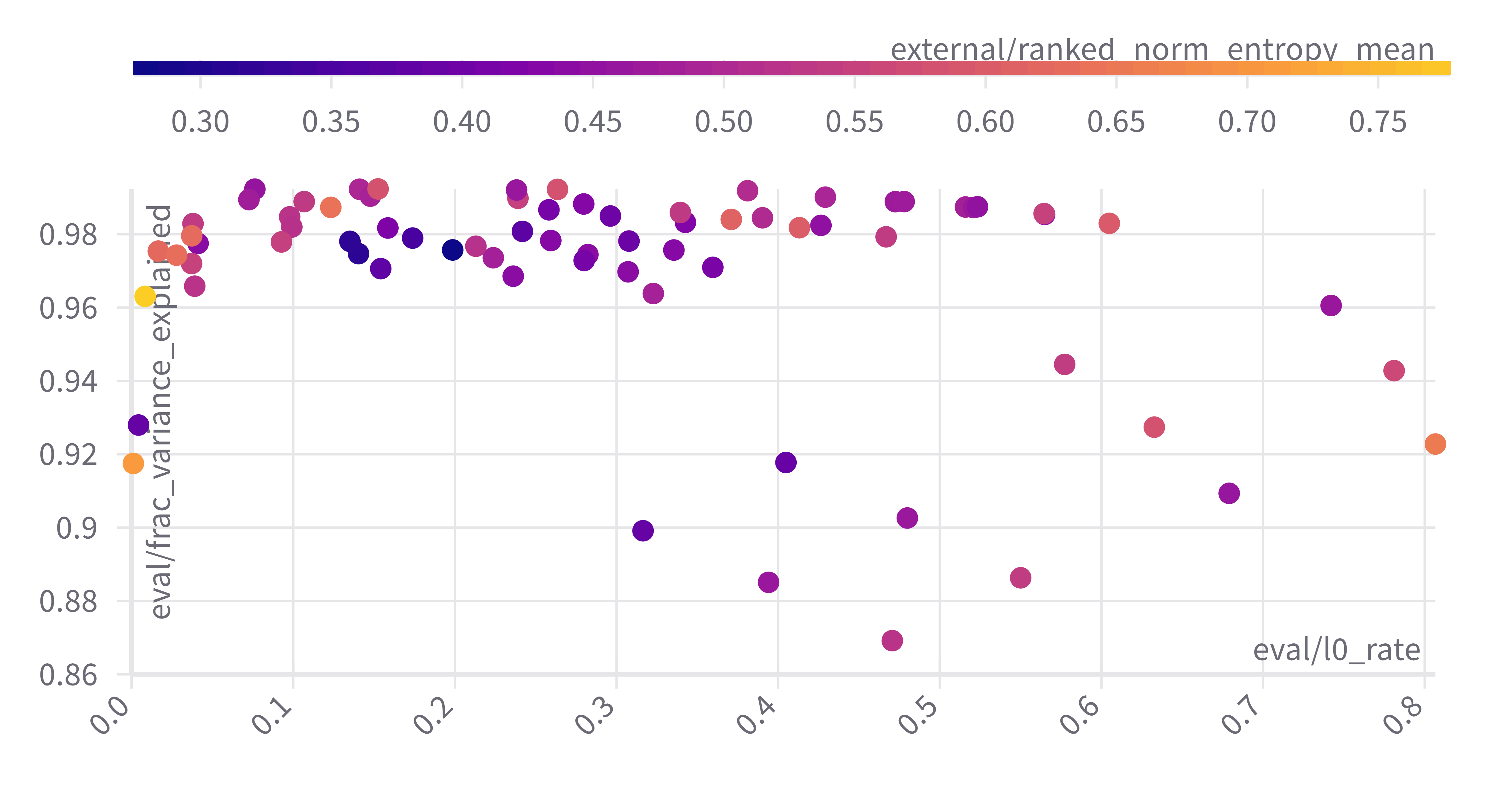}
    \caption{Mean entropy of top-correlated neurons across different sparsity–reconstruction trade-offs.}
    \label{fig:l0-rec-entro}
\end{figure}

In Figure~\ref{fig:l0-rec-entro}, we observe that lower entropy—indicating more class-specific neurons—is generally associated with better sparsity–reconstruction trade-offs. However, this relationship is not strictly monotonic. Past a certain sparsity level (i.e., low $L_0$), entropy begins to increase again, possibly due to the emergence of overly sparse and under-expressive representations.

These findings suggest that traditional SAE metrics provide useful but incomplete signals regarding interpretability. Optimizing for sparsity and reconstruction alone may help but does not guarantee the emergence of disentangled or semantically meaningful features.

\subsection{MedGemma naming and evaluating framework}
\label{appendix:medgemma}

This section provides detailed information on the implementation and the prompts used in the MedGemma naming framework. 

The MedGemma model utilized was the 4b-it version with 4-bit quantization, using checkpoints available on HuggingFace.
The naming scripts were computed using the CheXpert training dataset. For each neuron, the model was provided with the 15 most activating images and the 15 median activating images. The term "median" refers to ranking the images by activation magnitude and selecting 15 images starting from the median.
For the detection evaluation, 30 activating images were randomly selected from the set of activating images, along with 30 random non-activating images. In cases where there were insufficient images, sampling with replacement was employed. Activating samples were defined as those with an activation greater than \(1 \times 10^{-7}\).

Figures~\ref{fig:medgemma-naming-prompt} and \ref{fig:medgemma-detection-prompt} show the MedGemma prompt templates used in our work.

\begin{figure}[t]
\centering
\begin{tcolorbox}[prompt, title={MedGemma naming prompt}]
\textbf{Prompt:}
Below I provide you with a sequence of chest X-ray images. Your task is to identify and describe the shared concept among 
these images. The shared concept could be medical, such as a specific condition or abnormality, or it could relate to non-medical 
aspects like patterns, artifacts, or other visual features.\\

For example, you might identify a recurring visual pattern such as opacities in a specific lung region, consistent positioning 
of medical devices, or artifacts that appear across the images. The shared concept could also relate to image quality, 
patient positioning, or technical aspects of the X-rays.\\

Some concepts may be more abstract, such as patterns of disease progression, specific diagnostic features, or technical aspects like image clarity or exposure.\\

Further, some images may have been included that do not relate to the concept at all. Please ignore those items.\\

First, please summarize the main concept in a short sentence or phrase, approximately 5 to 10 words. Be as specific as possible-
it is fine if it doesn't fit all images. Please be precise and creative. Then, also provide a longer description elaborating 
on the concept.\\

Return the results in JSON format with the following JSON schema:\\
\texttt{\{\\
  "concept": "main concept summary",\\
  "description": "longer description of the shared concept"\\
}\}\\

The images follow now.
\end{tcolorbox}

\caption{Input naming prompt used by MedGemma.}
\label{fig:medgemma-naming-prompt}
\end{figure}

\begin{figure}[t]
\centering
\begin{tcolorbox}[prompt, title={MedGemma detection prompt}]
\textbf{Prompt:}
You will receive a specific concept and a description, such as "Pulmonary edema and pleural effusions in a patient with a central line" or "Technical Artifacts and/or Equipment-Related Anomalies."\\
Following this, you will be presented with multiple image examples.\\
Your objective is to assess which of these examples accurately contains the given concept.\\
For each image example provided, indicate with a 1 if the image is correctly labeled, or a 0 if it is mislabeled.\\
Ensure your response is formatted strictly as a valid Python list of length \{bs\}.
Return only the Python list without any additional information.\\

Concept: \{concept\}\\
Description: \{description\}\\
Image examples:
\end{tcolorbox}

\caption{Input detection prompt used by MedGemma.}
\label{fig:medgemma-detection-prompt}
\end{figure}

\end{document}